\documentclass{article}
\usepackage{spconf,amsmath,graphicx,hyperref}
\usepackage{booktabs}
\usepackage{multirow}
\usepackage{amsfonts} 

\title{ToPT: Task-Oriented Prompt Tuning for Urban Region Representation Learning}

\name{Zitao Guo$^{1,\dag}$, 
Changyang Jiang$^{2,\dag}$\thanks{$^\dag$ Equal contribution.}, 
Tianhong Zhao$^{2}$, 
Jinzhou Cao$^{2}$, 
Genan Dai$^{2,*}$, 
Bowen Zhang$^{2,*}$\thanks{$^*$ Corresponding authors}}
\address{$^{1}$College of Applied Science, Shenzhen University, Shenzhen, China \\
$^{2}$School of Artificial Intelligence, Shenzhen Technology University, Shenzhen, China
}
\begin{document}
\ninept 
\maketitle
\begin{abstract}
Learning effective region embeddings from heterogeneous urban data underpins key urban computing tasks (e.g., crime prediction, resource allocation). However, prevailing two-stage methods yield task-agnostic representations, decoupling them from downstream objectives. Recent prompt-based approaches attempt to fix this but introduce two challenges: they often lack explicit spatial priors, causing spatially incoherent inter-region modeling, and they lack robust mechanisms for explicit task–semantic alignment. We propose ToPT, a two-stage framework that delivers spatially consistent fusion and explicit task alignment. ToPT consists of two modules: spatial-aware region embedding learning (SREL) and task-aware prompting for region embeddings (Prompt4RE). SREL employs a Graphormer-based fusion module that injects spatial priors—distance and regional centrality—as learnable attention biases to capture coherent, interpretable inter-region interactions. Prompt4RE performs task-oriented prompting: a frozen multimodal large language model (MLLM) processes task-specific templates to obtain semantic vectors, which are aligned with region embeddings via multi-head cross-attention for stable task conditioning. Experiments across multiple tasks and cities show state-of-the-art performance, with improvements of up to 64.2\%, validating the necessity and complementarity of spatial priors and prompt–region alignment. 
The code is available at \url{https://github.com/townSeven/Prompt4RE.git}.

\end{abstract}

\section{Introduction}
\label{sec:intro}
Learning region embeddings is a fundamental task in urban computing. The aim is to encode heterogeneous urban data into compact vectors that downstream models can directly use, such as point-of-interests (POIs), human mobility, land use, satellite imagery, street-view images, and geo-text \cite{cao_UrbanMMCLUrbanRegion_2026,dai2025crrl,Urban2Vec}. 
These embeddings support key applications, including crime prediction, check-in forecasting, and service call estimation, and they can be reused across tasks and cities \cite{cao_UrbanRepresentationLearning_2025,MGFN,USPM}. 
With the rapid growth of publicly available urban data, region representation has become a unified interface between rich city data and decision models.

In recent years, deep learning has become the dominant approach for learning region representations.
Early studies mainly focused on integrating multi-view data such as mobility, POIs, and land use. 
They often produced a single region vector by simple operations, including feature concatenation, weighted summation, or compression with MLPs and autoencoders\cite{CGAL,MP-VN,ReMVC}. Later approaches highlighted the importance of spatial relations between regions. They introduced multi-graph structures and graph neural network (GNN) encoders\cite{Region2Vec,ROMER} to capture region–region interactions, and applied attentive fusion at the view level and sometimes also at the region level, as shown in models like MVURE \cite{MVURE},   HAFusion \cite{HAFusion} etc. Recent work extended region representation learning to multimodal settings. Beyond structured views, these methods incorporate satellite images, street-view data, and POI texts, and employ contrastive or hierarchical designs for cross-modal alignment, such as in UrbanMMCL\cite{cao_UrbanMMCLUrbanRegion_2026}, RegionDCL\cite{RegionDCL}, UrbanCLIP\cite{UrbanCLIP}, and CityFM\cite{CityFM}.


To further enhance task adaptation, recent studies have introduced prompt learning into region representation. These methods first obtain general-purpose region embeddings and then inject soft prompts or construct task-aware signals from multimodal inputs such as street-view images and geographic texts. 
In this way, prompts aim to align embeddings with the semantics of downstream tasks. 
For example, HREP \cite{HREP} applies randomly initialized soft prompts for parameter-efficient adaptation, while FlexiReg \cite{FlexiReg} leverages image and text prompts to enrich semantic information and improve generalization across tasks and cities. 
Despite these advances, existing approaches still face two major challenges.
(1) Current methods often ignore spatial relations across views. They usually encode each view independently, which captures intra-view dependencies but fails to model inter-view spatial interactions, leading to incomplete or inconsistent cross-view relations.
(2) Existing prompt mechanisms lack an explicit alignment with task objectives. As a result, task knowledge cannot be precisely mapped into region representations, thereby limiting predictive accuracy in downstream tasks.


To address the above issues, we propose \textbf{T}ask-\textbf{O}riented \textbf{P}rompt \textbf{T}uning for Urban Region Representation Learning (\textbf{ToPT}), a simple yet effective framework that enhances predictive accuracy by leveraging spatially consistent fusion and explicit task alignment.
ToPT consists of two modules: the spatial-aware region embedding learning (SREL) and the task-aware prompting module for region embeddings (Prompt4RE).
SREL learns task-agnostic region embeddings from multi-view urban data. We construct a Graphormer \cite{Graphormer}-based multi-view fusion mechanism that injects distance-based adjacency and node centrality as learnable attention biases, thereby capturing the interactions between different views.
Prompt4RE performs task-oriented prompting: frozen Multi-modal Large Language Models (MLLMs) process satellite images, street-view images, and geo-text under task-specific templates; we take the last-layer hidden states as prompt vectors and align them to region embeddings via multi-head cross-attention in a shared semantic space. The aligned prompts are projected as soft prompts and concatenated with region embeddings for downstream prediction. This design provides explicit spatial priors in fusion and explicit task alignment for prompts.



The main contributions of this paper can be summarized as follows:

\begin{itemize}
 
\item We propose ToPT, a simple yet effective two-stage framework for region representation learning that jointly enforces spatially informed fusion and explicit task alignment.

\item We develop Prompt4RE, a task-oriented prompting module that uses a frozen MLLM with task-specific templates to extract prompt vectors and aligns them to region embeddings via multi-head cross-attention, enabling explicit prompt–region alignment and stable task conditioning.

\item We provide comprehensive experiments across multiple downstream tasks and cities, demonstrating consistent state-of-the-art performance. ToPT achieves up to 64.2\% improvement over strong baselines. 

\end{itemize}

\begin{figure*}[t!]
    \centering
        \includegraphics[width=0.87\linewidth]{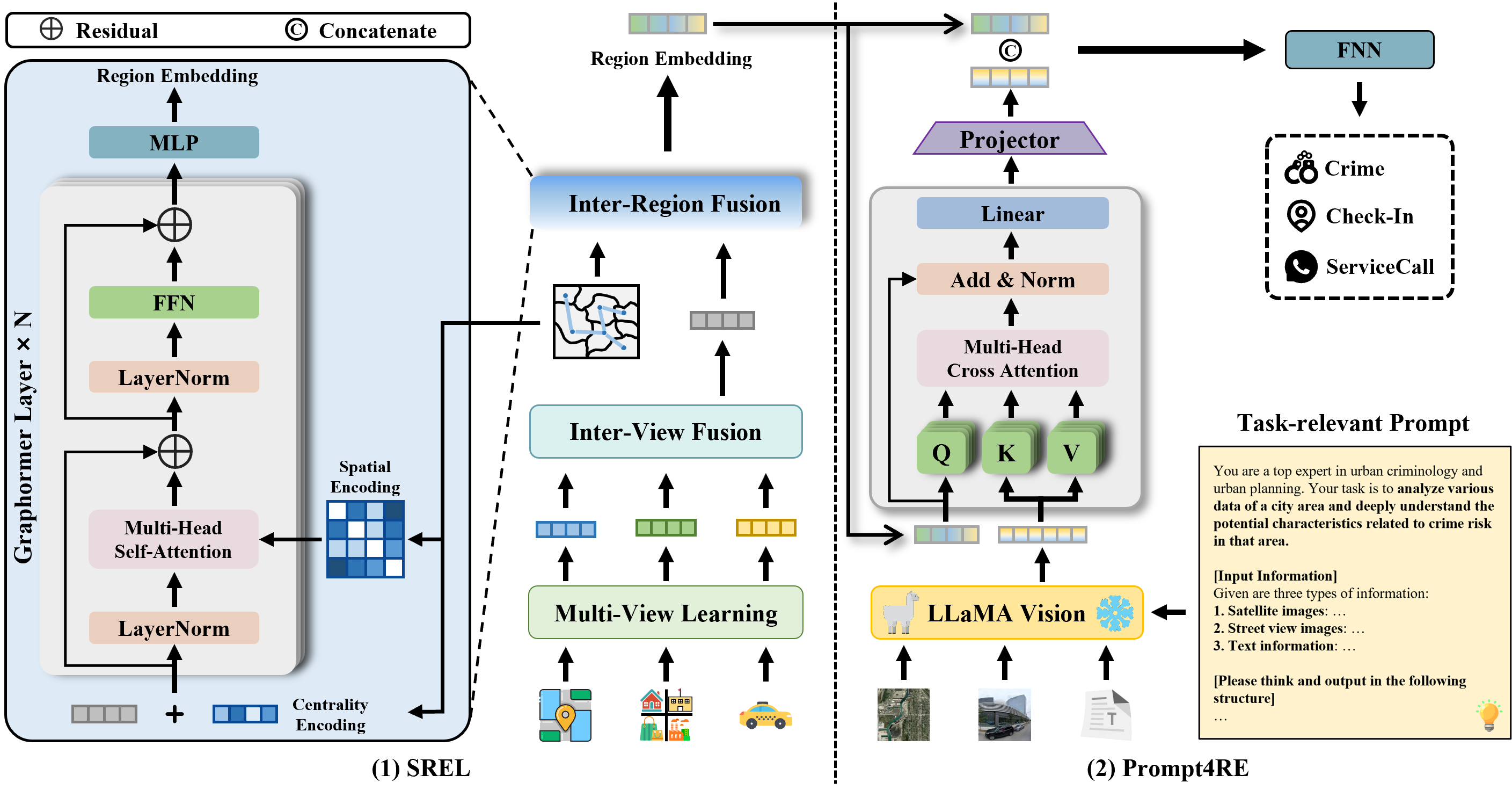}
    \caption{Our Model Framework}
    \label{fig:framework}
\end{figure*}

\section{Methodology}
\label{sec:majhead}

\subsection{Problem Definition}
\label{ssec:subhead}



Given a spatial area partitioned into $n$ non-overlapping regions with publicly accessible multi-view features, including POIs, mobility data, land-use categories, geographical neighbor information, satellite imagery, street-view images, and geo-text, we aim to learn a task-agnostic and transferable embedding function $f$ that maps each region $r_i$ to a $d$-dimensional vector $e_i \in \mathbb{R}^d$, such that $f: r_i \mapsto e_i$, forming the region embedding matrix $E = \{ e_1, \dots, e_n \}$. These embeddings are designed to be reusable across different downstream tasks, 
and can be directly used or further adapted via task-specific prompting for prediction in various urban computing scenarios.

\subsection{ToPT framework}
\label{ssec:model}

\subsubsection{Model Overview}

As illustrated in Fig. \ref{fig:framework}, the proposed ToPT framework comprises SREL and Prompt4RE.
SREL first encodes multi-view urban data into task-agnostic region representations with explicit spatial relationships. Subsequently, Prompt4RE takes these general-purpose embeddings and aligns them with task-specific context obtained from multi-modal urban prompts. 
Finally, the fused representation, which encapsulates both spatial and task-aware information, is passed through a fully-connected layer to generate the final predictions for various downstream urban computing tasks. This design ensures a coherent flow from general feature extraction and spatial fusion to task-specific adaptation and efficient prediction.

\subsubsection{Spatial-aware Region Embedding Learning Module (SREL)}


First, we adopt the multi-view learning module to obtain view-specific intermediate region embeddings. Given input feature matrices from POIs, land-use, and human mobility data, denoted as \( X_p, X_l, X_m \) respectively, the module produces corresponding embeddings \( Z_p, Z_l, Z_m \). 
It captures intra-view region correlations through efficient attention mechanisms, yielding informative view-specific representations.
Then, we develop an inter-view fusion stage to model cross-view interactions and adaptively estimate the importance of each view, combining them into a fused inter-view representation \( \tilde{H} \in \mathbb{R}^{N \times d} \).
Subsequently, to explicitly model the spatial correlations between regions, we propose a spatially-informed inter-region fusion module. 
This module takes \( \tilde{H} \) and a distance-based adjacency matrix \( A \in \mathbb{R}^{N \times N} \) as inputs. We enhance the input representations by incorporating structural information derived from the spatial graph. Specifically, for each region \( i \), we compute its strength centrality from the adjacency matrix \( A \) as \( c_i = \sum_j A_{ij} \), which reflects the region's overall connectivity. These values are max-normalized to ensure scale invariance, linearly projected to a \( d \)-dimensional space, and added to the original features \( \tilde{H} \). This results in a structurally-augmented feature matrix \( H' \), which encodes both the original semantic attributes and the structural importance of each region.

The core of this module is a multi-head self-attention mechanism that is augmented with spatial priors. The queries \( Q \) and keys \( K \) are obtained by applying linear projections to the enhanced features \( H' \). The attention weight between region \( i \) and region \( j \) is computed as:
\begin{equation}
    \alpha_{ij} = \text{Softmax}_j \left( \frac{(W_Q H'_i) (W_K H'_j)^\top}{\sqrt{d}} + B_{ij} \right),
\end{equation}
where \( W_Q, W_K \) are learnable projection matrices, and \( B_{ij} \) is a spatial bias term derived from the adjacency matrix \( A \). This formulation enables the attention mechanism to jointly consider feature similarity and spatial proximity, promoting spatially coherent representations. The model stacks \( L \) such layers, each followed by a position-wise feed-forward network with residual connections and layer normalization. The output is refined by a lightweight MLP, yielding the final spatially informed region embeddings \( E \in \mathbb{R}^{N \times d} \).
The module is optimized using multi-task learning objectives~\cite{HAFusion}. 

\subsubsection{Task-aware Prompting Module (Prompt4RE)}
Prompt4RE exploits the multimodal fusion and comprehension capacities of frozen MLLMs to derive task-specific contextual knowledge from diverse urban data.
Specifically, we collect satellite imagery, street-view photos, and geographic textual descriptions (following the FlexiReg template, which includes coordinates, addresses, and surrounding POI information). We design tailored prompt templates to guide the MLLM in extracting task-relevant information from these multimodal inputs; an example for crime prediction is shown in Fig.~\ref{fig:framework}. The last-layer hidden states of the MLLM are extracted as the prompt vectors  \( P \). 

To align these task-aware prompts with the spatial region embeddings \( E \in \mathbb{R}^{N \times d} \), we employ a multi-head cross-attention mechanism. The queries are derived from \( E \), while keys and values are projected from \( P \). The aligned representation is computed through multi-head cross-attention:
\begin{equation}
    \text{att}_i = \text{Softmax}\left( \frac{E W_Q^{(i)} (P W_K^{(i)})^\top}{\sqrt{d_i}} \right) P W_V^{(i)},
\end{equation}
where \( W_Q^{(i)}, W_K^{(i)}, W_V^{(i)} \) are learnable projection matrices for each head, and \( d_i \) is the dimension of each attention head. The outputs of all heads are concatenated:
\begin{equation}
    \text{MHA}(E, P) = \text{Concat}(\text{att}_1, \dots, \text{att}_L),
\end{equation}
where \( L \) is the number of attention heads.

To preserve the original spatial semantics while incorporating task-specific information, we add a residual connection followed by layer normalization, resulting in prompt-aligned region embeddings:
\begin{equation}
    P' = \text{LayerNorm}\left( \text{MHA}(E, P) + E W_{\text{res}} \right),
\end{equation}
where \( W_{\text{res}} \) are learnable projection matrices.  A linear layer then projects \( P' \) to match the dimensionality of \( E \), producing the soft prompts \( S \in \mathbb{R}^{N \times d} \). These are concatenated with the original region embeddings to form the final enhanced representations \( F = [E \| S] \in \mathbb{R}^{N \times 2d} \).

For each downstream prediction task, we employ a fully-connected layer that takes the enhanced representation \( f_i \) of region \( r_i \) as input and produces the prediction \( \hat{y}_i = \text{FNN}(f_i) \). The entire module is optimized using the standard Mean Squared Error loss. 
\section{Experiments}
\label{sec:experiments}

\subsection{Experimental Setups}

\textbf{Dataset.}
We use real-world data from the City of Chicago Data Portal\footnote{\url{https://data.cityofchicago.org/}} to train and evaluate our model.
Following \cite{HAFusion}, we adopt POIs, taxi trip records, and land-use categories to train the region representation module. 
To support the prompt learning module, we additionally construct multimodal inputs for each region: (i) satellite and street-view imagery retrieved via Google Maps\footnote{\url{https://www.google.com/maps}}; and (ii) geographic coordinates, addresses, and POIs obtained from OpenStreetMap\footnote{\url{https://www.openstreetmap.org/}}, which we further aggregate into geo-text descriptions. 


        

\begin{table*}[t!]
\caption{\textbf{Performance comparison on three downstream tasks.} The best result on each task is in \textbf{bold}. * indicates a statistically significant improvement with $p$-value $<$ 0.05.}
\label{tab:results}
\vspace{10pt} 
\centering
\setlength{\tabcolsep}{6pt} 
\small 
    \begin{tabular}{l|ccc|ccc|ccc}
        \toprule
            \multirow{2}{*}{\textbf{Chicago}} & \multicolumn{3}{c|}{\textbf{Crime}} & \multicolumn{3}{c|}{\textbf{Check-in}} & \multicolumn{3}{c}{\textbf{Service Call}} \\
            \cmidrule(lr){2-4} \cmidrule(lr){5-7} \cmidrule(lr){8-10}
            & \textbf{MAE $\downarrow$} & \textbf{RMSE $\downarrow$} & \textbf{\textit{R}$^{\textbf{2}}$ $\uparrow$} & \textbf{MAE $\downarrow$} & \textbf{RMSE $\downarrow$} & \textbf{\textit{R}$^{\textbf{2}}$ $\uparrow$} & \textbf{MAE $\downarrow$} & \textbf{RMSE $\downarrow$} & \textbf{\textit{R}$^{\textbf{2}}$ $\uparrow$} \\
        \midrule
            MVURE & 100.4 & 129.2 & 0.461 & 1693 & 3171 & 0.656 & 190.3 & 266.9 & 0.441 \\
            MGFN & 107.4 & 137.9 & 0.386 & 1281 & 2276 & 0.817 & 208.2 & 293.4 & 0.329 \\
            HREP & 88.3 & 114.4 & 0.578 & 1679 & 3135 & 0.664 & 185.7 & 262.2 & 0.468 \\
            ReCP & 86.9 & 120.1 & 0.534 & 1272 & 2341 & 0.804 & 206.7 & 303.4 & 0.284 \\
            RegionDCL & 121.7 & 159.6 & 0.179 & 2427 & 4184 & 0.402 & 195.7 & 272.1 & 0.445 \\
            HAFusion & 77.8 & 107.1 & 0.631 & 929 & 1947 & 0.870 & 159.3 & 222.0 & 0.613 \\
            FlexiReg & {61.7} & {85.1} & {0.766} & {922} & {1775} & {0.891} & {121.1} & {178.2} & {0.753} \\
        \midrule
            \textbf{ToPT*} & \textbf{49.3} & \textbf{62.7} & \textbf{0.874} & \textbf{424.6} & \textbf{634.6} & \textbf{0.986} & \textbf{95.6} & \textbf{127.4} & \textbf{0.874} \\
        \midrule
            w/o Prompt4RE & 62.5 & 82.4 & 0.781 & 753.7 & 1419.4 & 0.931 & 124.5 & 172.6 & 0.768 \\
            w/o Task-relevant Prompt & {52.2} & {66.0} & {0.858} & {520.8} & {911.5} & {0.972} & {105.3} & {146.5} & {0.833} \\
            w/o P-R Alignment & {56.4} & {77.8} & {0.805} & {679.7} & {1315.1} & {0.941} & {111.9} & {155.9} & {0.811} \\
        \midrule
            \textbf{Improvement} & \textbf{20.1\%} & \textbf{26.3\%} & \textbf{14.1\%} & \textbf{53.9\%} & \textbf{64.2\%} & \textbf{10.7\%} & \textbf{21.1\%} & \textbf{28.5\%} & \textbf{16.1\%} \\
        
        \bottomrule
    \end{tabular}
\end{table*}

\textbf{Baselines and Metrics.}
We compare our model against several representative baseline methods, including MVURE\cite{MVURE}, MGFN\cite{MGFN}, HREP\cite{HREP}, ReCP\cite{ReCP}, RegionDCL\cite{RegionDCL}, HAFusion\cite{HAFusion} and FlexiReg\cite{FlexiReg}. To evaluate model performance, following~\cite{HAFusion, FlexiReg}, we adopt three widely used metrics: Mean Absolute Error (MAE), Root Mean Squared Error (RMSE), and the coefficient of determination (R²).

\textbf{Model Parameters.}
In our experiments, we set both the region embeddings and soft prompts to a dimensionality of 144, with 4000 training epochs for the representation learning stage and 3000 epochs for the prompt learning stage. We adopted different settings for the three tasks. 
For crime prediction, we used a learning rate of 0.0007, 2 Graphormer layers, and a spatial bias weight of $\lambda$ = 0.1 in the representation learning stage, and a learning rate of 0.0001 in the prompt learning stage. For check-in prediction, the representation learning stage was trained with a learning rate of 0.0002, 5 Graphormer layers, and $\lambda$ = 0.4, while the prompt learning stage used a learning rate of 0.0003. 
For service call estimation, we set the representation learning stage with a learning rate of 0.0008, 2 Graphormer layers, and $\lambda$ = 0.3, followed by a prompt learning stage with a learning rate of 0.0001.
We run the method 5 times and report the average score for our method. Our implementation uses PyTorch 2.4.0 with Nvidia A100.

\subsection{Overall Performance}
\label{ssec: overall performances}

Table \ref{tab:results} presents the performance comparison of ToPT against state-of-the-art baselines on three representative tasks in Chicago: crime prediction, check-in prediction, and service call estimation. 
ToPT consistently outperforms all baselines by a significant margin, with an average improvement of 20.2\%, 42.9\%, and 21.9\% on the three tasks, respectively. We conduct $t$-tests against the strongest baseline and obtain $p$-values below 0.05 across all tasks, confirming the statistical significance of ToPT's performance gains.

Specifically, we find that the two-stage paradigms adopted by MVURE, MGFN, ReCP, RegionDCL, and HAFusion generally underperform, possibly due to the inability to flexibly align the learned embeddings with task objectives.
Furthermore, HREP attempts to address the limitations of two-stage methods by introducing randomly initialized soft prompts. However, the lack of semantic priors about urban regions prevents it from providing effective guidance during the prompt learning stage, resulting in suboptimal performance. FlexiReg, on the other hand, incorporates additional street-view images and geographic texts to enrich the understanding of region representations. Despite achieving certain performance improvements, its prompt learning remains task-agnostic and fails to capture the spatial autocorrelations among regions, which are crucial for urban pattern modeling.

In contrast, our ToPT overcomes these limitations by combining spatially-informed fusion and task-oriented prompt tuning. The former explicitly models inter-region correlations by injecting distance and centrality priors, while the latter aligns task-specific multimodal urban knowledge with region embeddings in a shared semantic space. This enables ToPT to generate more task-adaptive and semantically grounded representations, leading to robust performance gains across all benchmarks.

\subsection{Ablation Study}
\label{ssec: ablation study}

To further investigate the contribution of each component, we design three ablated variants: w/o Prompt4RE, which removes the entire prompt-tuning stage and directly uses the regional embeddings for downstream tasks; w/o Task prompt, which discards the task-specific templates; and w/o P–R Alignment, which replaces the alignment module with simple concatenation. As shown in Table \ref{tab:results}, removing Prompt4RE leads to a substantial performance drop, highlighting the critical role of the prompt-tuning stage in enhancing region embeddings. Interestingly, even without prompt learning, ToPT remains competitive with the strongest baseline FlexiReg, underscoring the effectiveness of our spatially-informed fusion. Furthermore, both task prompts and the P–R Alignment module prove beneficial: the templates guide MLLMs to extract task-relevant knowledge, while alignment bridges the semantic gap between prompt vectors and region embeddings.



\subsection{Generalizability Across Different MLLMs}
\label{ssec: generalizability_mllms}

\begin{figure}[h!]
    \centering
    \includegraphics[width=0.86\linewidth]{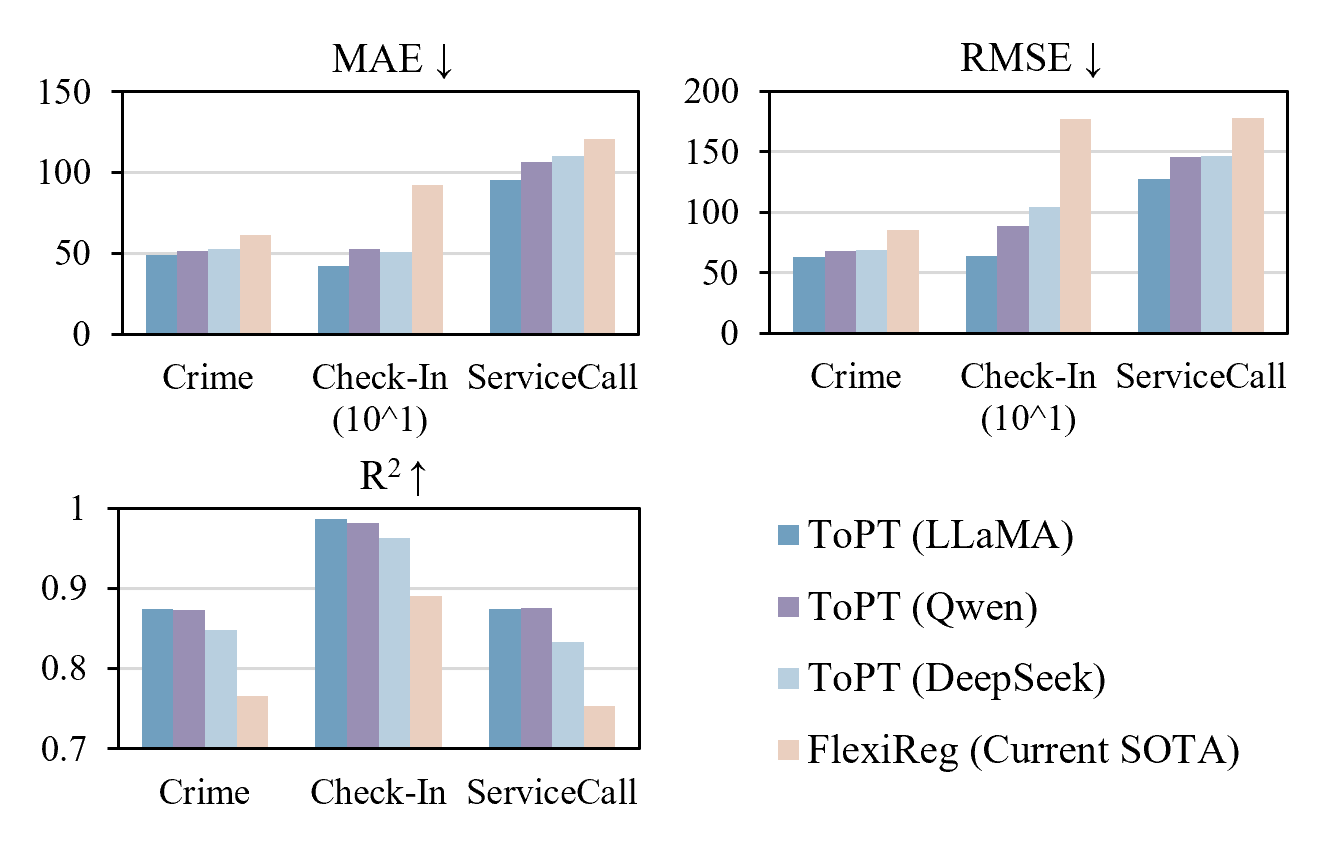}
    \caption{Impact of MLLMs.}
    \label{fig:chart}
\end{figure}

To demonstrate the generalizability of ToPT across different MLLMs, we compare the performance of three models: LLaMA-3.2-11B-Vision-Instruct, Qwen2.5-VL-7B-Instruct \cite{Qwen}, and DeepSeek-VL2 (4.5B) \cite{DeepSeek}. As shown in Fig. \ref{fig:chart}, ToPT consistently outperforms the FlexiReg baseline across all three tasks, regardless of the choice of MLLM. This indicates that the effectiveness of our approach does not rely on a specific large model but can generalize well to different MLLMs. The performance variations among the three models are relatively small, with LLaMA slightly ahead in the ServiceCall task. These results highlight the model-agnostic nature of ToPT and its ability to leverage various MLLMs to achieve strong performance in urban representation learning tasks.

\section{Conclusion}
\label{sec:conclusion}

In this paper, we introduced ToPT, a novel framework for urban region representation learning that enhances general region embeddings through task-oriented prompt tuning with MLLMs. ToPT consists of two key components: a spatial-aware region embedding module and a task-aware prompting module. The former leverages POIs, land-use categories, and human mobility features to derive robust region embeddings, with our Spatially-Informed Fusion module effectively capturing spatial autocorrelation during fusion. To further adapt these embeddings for downstream applications, we proposed Prompt4RE, a task-oriented prompt tuning mechanism that customizes task-specific templates to guide MLLMs in extracting relevant information. 
By aligning the resulting prompts with the region embeddings to form soft prompts, ToPT achieves strong adaptability across tasks. 
Extensive experiments demonstrate that ToPT consistently surpasses competitive baselines, achieving state-of-the-art performance with improvements of up to 64.2\%.

\vfill
\pagebreak
\section{Acknowledgments}
This research is supported by the Natural Science Foundation of Top Talent of SZTU (grant no.GDRC202518, no.GDRC202320), Shenzhen Science and Technology Program (No.RCBS20231211090548077, No.JCYJ20240813113218025, No.JCYJ20240813113300001), Project for Improving Scientific Research Capabilities of Key Construction Disciplines in Guangdong Province (2025ZDJS039).
\bibliographystyle{IEEEbib}
\bibliography{strings,refs}

\end{document}